\documentclass{article}
\usepackage[utf8]{inputenc}
\usepackage{amsmath}
\usepackage{newtxtext,newtxmath}
\usepackage{algorithmic}
\usepackage{algorithm}
\usepackage{graphicx}
\usepackage{textcomp}
\usepackage{booktabs}
\usepackage{multirow}
\usepackage{hyperref}
\hypersetup{hidelinks}
\usepackage{xcolor}
\usepackage[authoryear,round]{natbib}
\usepackage{placeins}
\usepackage{float}
\usepackage[letterpaper,left=1.5in,right=1.5in,top=1in,bottom=1in]{geometry}
\setcitestyle{authoryear,round}
\newcommand{\std}[1]{{\scriptsize(#1)}}
\makeatletter
\renewenvironment{abstract}{%
  \begin{center}
    {\large\scshape\MakeUppercase{\abstractname}\par}
  \end{center}
  \vspace{0.4em}
  \begin{quote}
}{%
  \end{quote}
  \vspace{0.6em}
}
\makeatother

\title{Progress- and Reliability-Oriented Group Policy Optimization\\for Agentic Reinforcement Learning}

\author{
\begin{tabular}{cc}
\begin{tabular}[t]{c}
Mingxuan Fan\\
{\small Baidu Inc., China}
\end{tabular}
&
\begin{tabular}[t]{c}
Peiyang Liu\\
{\small Peking University, China}
\end{tabular}
\end{tabular}
}
\date{}

\begin{document}
\maketitle

\begin{abstract}
Group-based reinforcement learning (RL) has become an effective paradigm for improving large language model agents on long-horizon interactive tasks. To obtain finer-grained policy updates than trajectory-level optimization, recent work has moved toward step-level group-based RL, where intermediate steps are grouped and compared within a rollout batch. However, step-level advantage estimation is sensitive to how groups are formed: grouping by broad state keys improves coverage but may compare actions taken under different histories, while enforcing historical consistency yields fairer comparisons at the cost of fragmented groups and missing peer-comparison signal. In this paper, we propose \textbf{ProGPO} (Progress- and Reliability-Oriented Group Policy Optimization), a learned-critic-free method for context-consistent step-level learning. ProGPO keeps exact-prefix action comparison, and complements sparse peer comparisons with transition credit derived from rollout-based state potentials. To estimate these potentials reliably, ProGPO combines semantic expansion with inverse-variance fusion across history depths. We evaluate ProGPO on two challenging agentic tasks, ALFWorld and WebShop, with Qwen2.5-1.5B-Instruct. Results show that ProGPO improves over matched agentic RL baselines under comparable computational overhead, and additional Qwen2.5-3B-Instruct experiments further test the scalability of the proposed method.
\end{abstract}

\section{Introduction}

Large language models (LLMs) are increasingly deployed as agents that perceive, reason, and act in external environments~\citep{brown2020language, yao2023react, liu2024agentbench, wang2024voyager}. Representative applications include web navigation~\citep{yao2022webshop, deng2023mind2web, zhou2023webarena, zheng2024seeact}, embodied household tasks~\citep{shridhar2021alfworld, ahn2022saycan, driess2023palme}, and tool-augmented reasoning~\citep{schick2023toolformer, qin2024toolllm, jin2025searchr1}. Unlike single-turn generation, these tasks require long-horizon planning, recovery from earlier mistakes, and credit assignment under sparse delayed rewards.

Reinforcement learning (RL) has therefore become a key post-training paradigm for improving model behavior, from human-preference optimization~\citep{christiano2017deep, ouyang2022training, bai2022training, rafailov2023direct} to reasoning-oriented RL~\citep{guo2025deepseek}. In particular, group-based methods estimate advantages from multiple sampled responses or trajectories for the same task, avoiding a learned critic while retaining scalable policy optimization. GRPO~\citep{shao2024deepseekmath} and variants such as DAPO~\citep{yu2025dapo} compare trajectory-level outcomes within a group, while RLOO~\citep{kool2019rloo} uses leave-one-out baselines. However, these methods are primarily designed for single-turn tasks such as mathematical reasoning and code generation, where each response receives a complete outcome and step-level credit is less ambiguous.

In multi-turn agentic tasks, direct trajectory-wise optimization becomes inefficient. Approaches such as RAGEN~\citep{wang2025ragen} and Search-R1~\citep{jin2025searchr1} concatenate the full interaction history into a single sequence, causing the effective context length to grow rapidly with the number of turns. To avoid this context explosion, recent step-level methods such as GiGPO~\citep{feng2025gigpo} group repeated intermediate states and compute within-group advantages, enabling finer-grained updates without per-step extra rollouts.

\begin{figure}[H]
\centering
\includegraphics[width=0.98\textwidth]{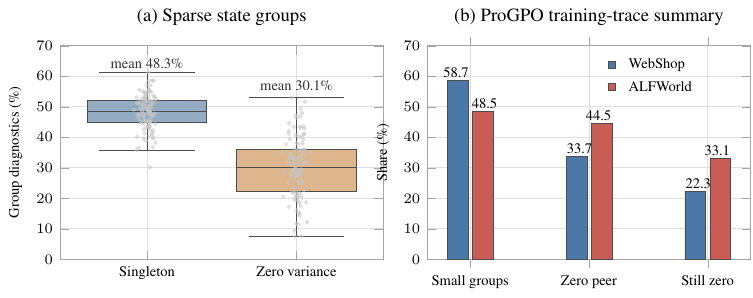}
\caption{Sparse-signal diagnostic. (a) WebShop exact state groups. (b) Training-trace shares for small value groups, zero peer signal, and residual zero-signal steps.}
\label{fig:sparse_signal}
\end{figure}

Nevertheless, step-level grouping introduces a new comparability requirement. If steps are grouped only by the current observation, then actions taken under different histories, goals, or available affordances may be compared in the same group, producing biased action advantages. A natural remedy is \emph{context consistency}: compare only steps that share the same historical prefix. Under this stricter grouping, however, the batch becomes fragmented. Many steps become \emph{singletons} or low-contrast groups, leaving no usable peer-comparison signal. Figure~\ref{fig:sparse_signal} illustrates this phenomenon in ProGPO training traces: 33.7\% of WebShop steps and 44.5\% of ALFWorld steps have zero prefix-consistent peer-comparison advantage before transition credit is applied, while small value groups are common throughout training.

\begin{figure}[t]
\centering
\includegraphics[width=\textwidth]{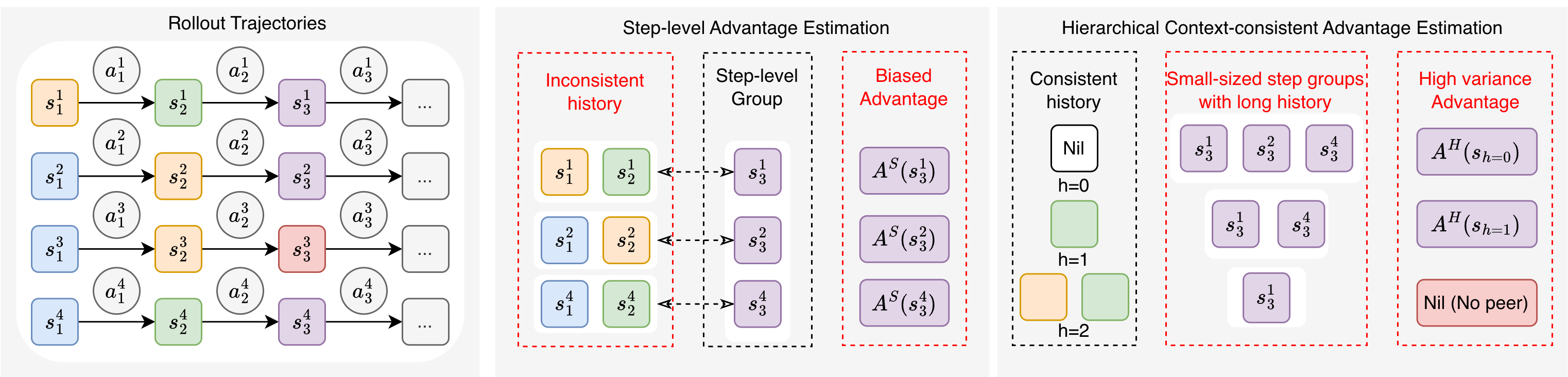}
\caption{Motivation for ProGPO. Left: rollout trajectories. Middle: step-level grouping by the current state can mix different histories. Right: context-consistent grouping creates smaller peer groups.}
\label{fig:motivation}
\end{figure}

To address this sparse-signal regime, we propose \textbf{ProGPO} (Progress- and Reliability-Oriented Group Policy Optimization), a step-level learning framework that adds progress-sensitive credit to context-consistent training without training a parametric critic or collecting additional rollouts. ProGPO separates action comparison from progress estimation: prefix-consistent groups provide action-relative advantages, and rollout-derived state potentials define transition credit $V(s_{t+1}) - V(s_t)$. Thus, action comparison remains exact, while state progression can still be learned from batch-level rollout statistics.

Estimating these state potentials is itself a reliability problem. Shallow groups contain more samples but blur history-dependent value; deeper prefixes are more specific but noisier. ProGPO therefore treats value lookup as a multi-resolution estimation problem and fuses eligible depths with inverse-variance weights based on sample size and return variance. Semantic expansion further stabilizes potential estimation, while action comparison remains exact. This design keeps the method free of a trained value network while supplying progress credit for steps that peer comparison cannot score.

We evaluate ProGPO on WebShop and ALFWorld using Qwen2.5-1.5B-Instruct. ProGPO improves the best matched ALFWorld baseline from 87.8\% to 90.1\% success and increases WebShop success from 67.6\% to 71.5\%. Additional Qwen2.5-3B-Instruct ALFWorld runs show the same late-stage advantage over GiGPO and HGPO in validation success and test score.

The main contribution of this work is a credit assignment framework that keeps the fairness of history-consistent comparison while recovering denser learning signals from the same rollouts. Rather than relaxing the grouping rule and risking unfair action comparisons, ProGPO keeps action comparison prefix-consistent and adds a separate transition-credit signal from rollout-derived state potentials. We further show how to estimate these potentials without training a parametric critic: semantic expansion supplies nearby evidence when exact groups are small, and inverse-variance multi-resolution fusion lets better-supported history depths carry more weight.

\section{Related Work}

\textbf{Reinforcement learning for LLMs.} RL has become a central ingredient in aligning and improving LLMs beyond supervised fine-tuning. PPO~\citep{schulman2017proximal} with a learned value function was standard in early RLHF pipelines~\citep{christiano2017deep, ouyang2022training, bai2022training}, while DPO~\citep{rafailov2023direct} showed that some preference-learning objectives can avoid explicit online RL. Recent reasoning models again rely heavily on RL-style post-training~\citep{guo2025deepseek}. GRPO~\citep{shao2024deepseekmath} removes the critic by comparing multiple sampled responses for the same prompt and computing group-relative advantages. Later refinements include DAPO~\citep{yu2025dapo} and RLOO~\citep{kool2019rloo}. These methods inherit the policy-gradient lineage of REINFORCE, actor-critic learning, and advantage-weighted updates~\citep{williams1992simple, sutton1999policy, konda1999actorcritic, peng2019advantage}, but are particularly effective on single-turn tasks such as mathematical reasoning and code generation, where one response constitutes the entire interaction. Step-level supervision has also been studied through learned process verifiers~\citep{cobbe2021training, lightman2024verify}; ProGPO instead derives step signals from rollout groups without human process labels or an additional verifier.

\textbf{Multi-turn LLM agents.} Deploying LLMs as interactive agents in sequential environments~\citep{yao2023react, shinn2023reflexion} raises challenges absent from single-turn settings, including long-horizon credit assignment, context management across turns, and non-stationary interaction dynamics. Early prompt-based approaches such as ReAct~\citep{yao2023react} and Reflexion~\citep{shinn2023reflexion} rely on in-context learning without parameter updates, but their performance often saturates on harder tasks. The benchmark landscape has expanded from simulated shopping and household environments~\citep{yao2022webshop, shridhar2021alfworld} to realistic web, GUI, and tool-use settings~\citep{deng2023mind2web, zhou2023webarena, koh2024visualwebarena, qin2024toolllm}. Related embodied and digital-agent systems ground LMs in robot or interface actions~\citep{ahn2022saycan, driess2023palme, brohan2023rt2, furuta2024multimodal, bai2024digirl}. More recent work applies RL to adapt agent policies through interaction, with substantial gains on web navigation, embodied tasks, and tool-augmented search~\citep{jin2025searchr1, wang2024voyager}.

\textbf{RL for long-horizon agents.} Applying group-based RL to multi-turn agents requires finer credit assignment across steps. Trajectory-wise approaches~\citep{wang2025ragen, jin2025searchr1} concatenate the full history into a single sequence but suffer from context explosion as episodes lengthen. GiGPO~\citep{feng2025gigpo} instead constructs step-level groups from repeated anchor states, while HGPO~\citep{zhang2025hgpo} improves fairness through hierarchical historical-context matching. ProGPO derives progress credit from rollout statistics without fitting a separate value network.

\section{Preliminaries}
\label{sec:prelim}

\subsection{Problem Definition}

We consider long-horizon agentic tasks where an LLM agent $\pi_\theta$ interacts with an environment over $T$ steps to accomplish a goal described by a natural language instruction $\boldsymbol{x}$. At each step $t$, the agent observes state $\boldsymbol{s}_t \in \mathcal{S}$ (comprising the environment observation and a bounded history window) and produces an action $\boldsymbol{a}_t \in \mathcal{V}^*$ as a variable-length token sequence. The environment transitions to a new state $\boldsymbol{s}_{t+1}$ based on the action. A sparse scalar reward $r \in \mathbb{R}$ is provided only upon episode termination, indicating task success or failure. We denote a full trajectory as $\tau = \{(\boldsymbol{s}_1, \boldsymbol{a}_1), \ldots, (\boldsymbol{s}_T, \boldsymbol{a}_T), r\}$, and define the discounted step return as $R_t = \gamma^{T-t} r$, where $\gamma \in (0, 1]$ controls the temporal discount. The objective is to maximize expected return $\mathbb{E}_{\tau \sim \pi_\theta}[r]$.

\subsection{Group-Based RL for Agents}

Group-based methods estimate advantages without a learned critic by sampling multiple rollouts for the same task and comparing their outcomes. Given $N$ trajectories $\{\tau^{(1)}, \ldots, \tau^{(N)}\}$ for task $\boldsymbol{x}$, the trajectory-level group advantage (as in GRPO) is:
\begin{equation}
    A^{\text{traj}}(\tau^{(i)}) = \frac{r^{(i)} - \mu}{\sigma}, \quad \mu = \frac{1}{N}\sum_{j=1}^N r^{(j)}, \quad \sigma = \text{std}(\{r^{(j)}\}_{j=1}^N).
\end{equation}
This advantage is shared across all tokens in the trajectory, providing a coarse learning signal that does not distinguish between steps that contributed to success and those that did not.

\subsection{Trajectory-Level vs.\ Step-Level Advantage}

To achieve finer credit assignment, stepwise frameworks~\citep{feng2025gigpo, zhang2025hgpo} decompose trajectories into individual steps and estimate advantages at the step level. Under this framework, the policy is conditioned on a bounded context: $\pi_\theta(\boldsymbol{a}_t | \boldsymbol{s}_t, \boldsymbol{h}_{t-K:t})$, where $\boldsymbol{h}_{t-K:t}$ represents a history window of length $K \ll T$. Steps are grouped by their observation context, and within-group comparison yields step-level advantages. For step $i$ with context $\bar{s}_i$, let $G_{\bar{s}_i} = \{j : \bar{s}_j = \bar{s}_i\}$ denote the set of steps sharing the same context. Following common notation for step-level group advantages, we use the superscript $S$ to denote ``step''. The grouped step-level advantage is:
\begin{equation}
    A^S(\bar{s}_i) =
    \frac{R_i - \frac{1}{|G_{\bar{s}_i}|} \sum_{j \in G_{\bar{s}_i}} R_j}
    {\sigma_{G_{\bar{s}_i}} + \epsilon},
    \label{eq:step_adv}
\end{equation}
where $\sigma_{G_{\bar{s}_i}}$ is the within-group standard deviation. This estimator isolates the effect of action choice by controlling for state, providing much richer feedback than trajectory-level advantages. However, it requires $|G_{\bar{s}_i}| \geq 2$ and non-identical returns within the group; when a step's context is unique within the batch (a \emph{singleton}) or all peers have the same return, no peer comparison is possible and the step receives zero learning signal.

\section{Proposed Method}
\label{sec:method}

\subsection{Limitations of Existing Stepwise Group-Based RL}
\label{sec:analysis}

We first analyze why stepwise group-based RL loses signal after enforcing context consistency.

The step-level advantage in Eq.~\eqref{eq:step_adv} requires at least two comparable steps with non-identical outcomes. In diverse environments with large state spaces, exact matching often produces \emph{singleton groups}, where no peer exists, or low-contrast groups whose returns are identical. In ProGPO training traces, zero prefix-consistent peer-comparison advantage occurs for 33.7\% of WebShop steps and 44.5\% of ALFWorld steps before transition credit is applied. Pure peer comparison therefore underuses many steps, including decisions that may affect later task progress.

Relaxing the grouping criterion can reintroduce comparisons across different goals, constraints, or available affordances. ProGPO keeps action comparison exact and introduces progress credit through state-potential estimation.

\FloatBarrier
\subsection{ProGPO: Progress- and Reliability-Oriented Group Policy Optimization}

\begin{figure}[t]
\centering
\includegraphics[width=\textwidth]{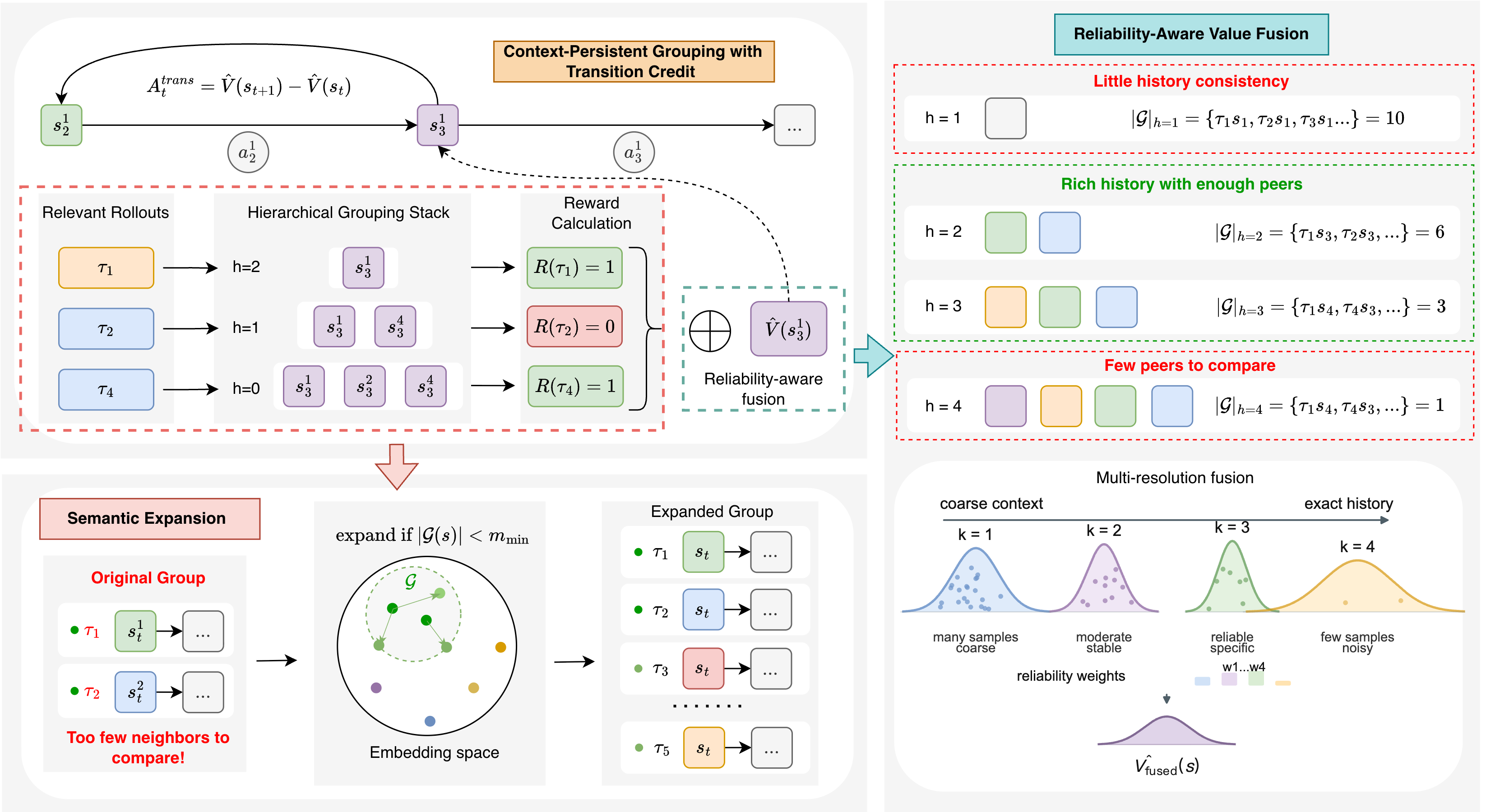}
\caption{Overview of ProGPO. State potentials are estimated from rollout outcomes, expanded with nearby states when needed, and fused across history depths before transition credit is computed.}
\label{fig:overview}
\end{figure}

We now describe ProGPO, a learned-critic-free credit assignment framework for context-consistent step-level RL (Figure~\ref{fig:overview}). The method keeps exact-prefix grouping for action-relative comparison. The remaining steps are handled through non-parametric state potentials estimated from rollout outcomes, with semantic expansion and reliability-aware multi-resolution fusion used to make these estimates less brittle. Thus ProGPO still performs value estimation, but it does so through rollout-batch statistics rather than a learned value model.

\subsubsection{Prefix-Consistent Step Comparison}
\label{sec:hierarchical}

At step $t$, let $\boldsymbol{o}_t$ denote the current environment observation and let
$\boldsymbol{p}^{(k)}_t=(\boldsymbol{o}_{t-k+1},\ldots,\boldsymbol{o}_{t})$ denote an exact observation prefix of depth $k$, where $k=1$ uses only the current observation and larger $k$ includes a longer history. Following the principle of context-consistent comparison~\citep{zhang2025hgpo}, ProGPO forms exact-prefix groups:
\begin{equation}
    G^{(k)}(\boldsymbol{p}^{(k)}_t)=\{j: \boldsymbol{p}^{(k)}_j=\boldsymbol{p}^{(k)}_t\}.
    \label{eq:hier_group}
\end{equation}
For each valid group, ProGPO computes a group-relative advantage:
\begin{equation}
    A^{(k)}_t =
    \frac{R_t - \mu(G^{(k)}(\boldsymbol{p}^{(k)}_t))}
    {\sigma(G^{(k)}(\boldsymbol{p}^{(k)}_t)) + \epsilon}.
    \label{eq:hier_adv}
\end{equation}
Groups with a single member or zero return variance contribute no action-comparison signal. The hierarchical advantage aggregates all valid depths:
\begin{equation}
    A^{\mathrm{hier}}_t =
    \frac{\sum_{k \in \mathcal{K}_t} (k+1)^{\alpha_h} A^{(k)}_t}
    {\sum_{k \in \mathcal{K}_t} (k+1)^{\alpha_h}},
    \label{eq:hier_agg}
\end{equation}
where $\mathcal{K}_t$ is the set of depths that produce a valid group-relative estimate. This estimator preserves prefix consistency while using all valid history depths.

\subsubsection{State Transition Scoring}
\label{sec:transition}

To score singleton and low-contrast steps, ProGPO introduces a non-parametric state potential and derives credit from state transitions. The prefix hierarchy supplies multiple estimates of $\hat{V}(s)$. Short prefixes often have enough samples to be stable, but they can blur history-dependent value. Long prefixes match the interaction context more closely, but the corresponding groups are often small. Selecting one depth globally is therefore brittle.

\textbf{Depth-specific state potentials.} For each eligible prefix group $G^{(k)}(\boldsymbol{p}^{(k)}_t)$ in a training batch, ProGPO estimates a depth-specific state potential as the mean return of steps sharing that prefix:
\begin{equation}
    \hat{V}^{(k)}(\boldsymbol{p}^{(k)}_t) =
    \frac{1}{|G^{(k)}(\boldsymbol{p}^{(k)}_t)|}
    \sum_{j \in G^{(k)}(\boldsymbol{p}^{(k)}_t)} R_j .
    \label{eq:value_table}
\end{equation}
This estimate is computed directly from rollout-batch statistics.

\textbf{Reliability-aware multi-resolution fusion.} ProGPO treats the hierarchy as a family of value estimators rather than selecting one depth. For the current state lookup, let $n_k$ and $\sigma_k^2$ be the sample count and empirical return variance of the eligible group at depth $k$. A depth participates only when $n_k$ exceeds a small minimum threshold. ProGPO assigns the precision-style reliability weight
\begin{equation}
    w_k = \frac{n_k}{\sigma_k^2 + \epsilon_{\mathrm{var}}},
    \label{eq:mr_weight}
\end{equation}
where $\epsilon_{\mathrm{var}}>0$ is a small variance floor that prevents unstable weights when empirical return variance is near zero. ProGPO then computes the fused potential
\begin{equation}
    \hat{V}^{\mathrm{mr}}(s_t)
    =
    \frac{\sum_{k \in \mathcal{M}_t} w_k \hat{V}^{(k)}(\boldsymbol{p}^{(k)}_t)}
    {\sum_{k \in \mathcal{M}_t} w_k},
    \label{eq:mr_value}
\end{equation}
where $\mathcal{M}_t$ contains the eligible depths for the state lookup. Equation~\eqref{eq:mr_value} gives larger influence to well-sampled, low-variance estimates.

Unlike the depth-only weighting in Eq.~\eqref{eq:hier_agg}, this fusion is data-dependent: two groups at the same depth can receive different weights when their sample counts or return variance differ.

\textbf{Transition credit.} Given the state potentials, we derive a directional learning signal for each step from the state transition:
\begin{equation}
    A^{\text{trans}}_t = \hat{V}^{\mathrm{mr}}(s_{t+1}) - \hat{V}^{\mathrm{mr}}(s_t),
    \label{eq:transition}
\end{equation}
which measures whether the action at step $t$ moves the agent toward a higher-potential state. This yields a progress signal without training a parametric critic.

\textbf{Semantic similarity expansion for potential stabilization.} When the $k=1$ observation group is small, ProGPO expands the value lookup with semantically similar states. In our implementation, the state embedding is the last-layer policy hidden state collected during policy evaluation at the action boundary. For a group with $|G_{\bar{s}}| < m$, we retrieve the top-$K_{\mathrm{sem}}$ nearest states and compute a weighted potential:
\begin{equation}
    \hat{V}^{\text{sem}}(\bar{s}) = \frac{\sum_{j \in G_{\bar{s}}} R(\bar{s}_j) + \sum_{\ell=1}^{K_{\mathrm{sem}}} \omega_\ell \cdot R(\bar{s}_{q_\ell})}{|G_{\bar{s}}| + \sum_{\ell=1}^{K_{\mathrm{sem}}} \omega_\ell},
    \label{eq:semantic_value}
\end{equation}
where $q_\ell$ indexes the $\ell$-th nearest neighbor by cosine similarity of hidden state embeddings, and $\omega_\ell = \text{softmax}(\text{sim}(\boldsymbol{h}_{\bar{s}}, \boldsymbol{h}_{q_\ell}) / \tau)$ are similarity-based weights.

\textbf{Use in state-potential estimation.} ProGPO uses semantic expansion only for state potentials; action advantages remain tied to exact-prefix comparison. This keeps action comparison prefix-consistent, although the final advantage can still be influenced by the expanded potential through transition credit.

\subsubsection{Complementary Credit Composition}
\label{sec:progpo_advantage}

ProGPO combines prefix-consistent peer comparison with transition credit. We set $A^{\mathrm{hier}}_t=0$ when no valid exact-prefix comparison is available. Let $c_t$ indicate whether peer comparison provides a nonzero signal, and let $\eta_t$ denote the residual gate applied to transition credit:
\begin{align}
    c_t &= \mathbf{1}\{|A^{\mathrm{hier}}_t|>\epsilon\}, \qquad
    \eta_t = 1 - (1-\lambda_{\mathrm{fb}})c_t, \label{eq:residual_gate}\\
    A^{\mathrm{ProGPO}}_t
    &= A^{\mathrm{hier}}_t
    + \lambda_{\mathrm{tr}} \eta_t A^{\mathrm{trans}}_t .
    \label{eq:final_adv}
\end{align}
Here $\mathbf{1}\{\cdot\}$ is the indicator function, $\lambda_{\mathrm{tr}}$ scales transition credit, and $\lambda_{\mathrm{fb}}$ controls how much transition credit remains when peer comparison is already active.

\subsection{The Policy Optimization Objective}

ProGPO optimizes the standard clipped surrogate:
\begin{align}
    \mathcal{J}_{\text{ProGPO}}(\theta)
    = \mathbb{E}\Bigg[
    \frac{1}{NT} \sum_{i=1}^{N} \sum_{t=1}^{T}
    \min\Big(
    &\rho_\theta(\boldsymbol{a}_t^{(i)}) A_t^{\text{ProGPO},(i)}, \nonumber \\
    &\text{clip}(\rho_\theta, 1\pm\epsilon) A_t^{\text{ProGPO},(i)}
    \Big)
    \Bigg]
    - \lambda \mathbb{D}_{\text{KL}}[\pi_\theta \| \pi_{\text{ref}}],
    \label{eq:objective}
\end{align}
where $\rho_\theta = \pi_\theta(\boldsymbol{a}_t | \boldsymbol{s}_t, \boldsymbol{x}) / \pi_{\text{old}}(\boldsymbol{a}_t | \boldsymbol{s}_t, \boldsymbol{x})$ is the importance sampling ratio and $\lambda$ controls the KL penalty.

\subsection{Analysis}

\textbf{Special cases.} Several familiar variants appear as ablations of ProGPO. Setting $\lambda_{\mathrm{tr}}=0$ removes transition credit, using only $k=1$ reduces the grouping to current-observation matching, disabling multi-resolution fusion gives a single-depth potential estimator, and setting $m=0$ removes semantic expansion.

\textbf{Computational overhead.} ProGPO computes group statistics and state potentials from the existing rollout batch. Multi-resolution fusion adds per-group variance estimates, and semantic expansion reuses hidden states from policy evaluation. No additional rollouts, parametric critic networks, or reward models are required.

\section{Experiments}
\label{sec:exp}

\subsection{Experiment Setup}

\textbf{Agentic benchmarks.} We evaluate on WebShop~\citep{yao2022webshop} and ALFWorld~\citep{shridhar2021alfworld}, two standard long-horizon agent benchmarks. WebShop requires agents to search for and purchase products matching natural-language instructions, with episodes lasting up to 30 steps. ALFWorld covers embodied household tasks such as finding objects, heating food, and cleaning items, with episodes lasting up to 50 steps.

\textbf{Models.} The main matched validation results use Qwen2.5-1.5B-Instruct. To test whether the training trend persists at a larger policy scale, we additionally run ALFWorld training-side comparisons with Qwen2.5-3B-Instruct.

\textbf{Compared methods.} For the 1.5B validation setting, we compare matched runs of GRPO, GiGPO, HGPO, and ProGPO, with three validation seeds for each method at step 160. The 3B ALFWorld scaling study compares GiGPO, HGPO, and ProGPO using training success, validation success, and validation test score. For broader context, Table~\ref{tab:main_results} also includes aggregate numbers for prompting, PPO, and RLOO from prior reports; conclusions about relative improvement are drawn from the matched local reruns.

\textbf{Reference methods.} GRPO~\citep{shao2024deepseekmath} provides trajectory-level group-relative advantages. GiGPO~\citep{feng2025gigpo} and HGPO~\citep{zhang2025hgpo} are step-level group-based methods for long-horizon agents. PPO~\citep{schulman2017proximal} and RLOO~\citep{kool2019rloo} are included through cited aggregate results.

\textbf{Training details.} The matched 1.5B runs use batch size 16, rollout group size 8, learning rate $1 \times 10^{-6}$, KL coefficient $0.01$, and vLLM rollout generation. WebShop episodes last up to 30 steps and ALFWorld episodes last up to 50 steps. ProGPO uses history length $K_h=2$, length-weight exponent $\alpha_h=1.0$, minimum value group size $m=3$, semantic neighbors $K_{\mathrm{sem}}=5$, transition scale $\lambda_{\mathrm{tr}}=0.5$, fallback scale $\lambda_{\mathrm{fb}}=0.3$, inverse-variance fusion with minimum multi-resolution group size 2, and variance floor $10^{-3}$.

\FloatBarrier
\subsection{Main Results}

\begin{table}[!htbp]
\centering
\caption{Performance on ALFWorld and WebShop. Prompting, PPO, and RLOO numbers are cited from prior reports; GRPO, GiGPO, HGPO, and ProGPO are local Qwen2.5-1.5B-Instruct reruns. Parentheses denote standard deviations over three seeds when available.}
\label{tab:main_results}
\vspace{0.45em}
\footnotesize
\setlength{\tabcolsep}{1.0pt}
\renewcommand{\arraystretch}{1.24}
\begin{tabular*}{\linewidth}{@{}ll@{\extracolsep{\fill}}ccccccc|cc@{}}
\toprule
Type & Method & \multicolumn{7}{c|}{\textbf{ALFWorld}} & \multicolumn{2}{c}{\textbf{WebShop}} \\
\cmidrule(lr){3-9}\cmidrule(lr){10-11}
& & Pick & Clean & Cool & Look & Heat & Pick2 & All & Score & Succ. \\
\midrule
\multicolumn{11}{@{}l}{\emph{Prompt-based methods}} \\
Prompt. & GPT-4o & 75.3 & 60.8 & 31.2 & 56.7 & 21.6 & 49.8 & 48.0 & 31.8 & 23.7 \\
Prompt. & Gemini-2.5-Pro & 92.8 & 63.3 & 62.1 & 69.0 & 26.6 & 58.7 & 60.3 & 42.5 & 35.9 \\
Prompt. & Qwen2.5 & 5.9 & 5.5 & 3.3 & 9.7 & 4.2 & 0.0 & 4.1 & 23.1 & 5.2 \\
Prompt. & ReAct & 17.4 & 20.5 & 15.7 & 6.2 & 7.7 & 2.0 & 12.8 & 40.1 & 11.3 \\
Prompt. & Reflexion & 35.3 & 22.2 & 21.7 & 13.6 & 19.4 & 3.7 & 21.8 & 55.8 & 21.9 \\
\midrule
\multicolumn{11}{@{}l}{\emph{RL-based methods}} \\
RL & PPO & 64.8\std{3.5} & 40.5\std{6.9} & 57.1\std{4.9} & 60.6\std{6.6} & 46.4\std{4.0} & 47.4\std{1.9} & 54.4\std{3.1} & 73.8\std{3.0} & 51.5\std{2.9} \\
RL & RLOO & 88.3\std{3.0} & 52.8\std{8.6} & 71.0\std{5.9} & 62.8\std{8.7} & 66.4\std{5.5} & 56.9\std{4.7} & 69.7\std{2.5} & 73.9\std{5.6} & 52.1\std{6.7} \\
RL & GRPO & 68.5\std{8.5} & 79.1\std{2.0} & 80.5\std{2.9} & 83.4\std{7.5} & 80.2\std{4.2} & 60.1\std{4.0} & 75.8\std{1.0} & 81.9\std{1.2} & 56.8\std{2.3} \\
RL & GiGPO & 73.9\std{7.6} & 85.6\std{3.0} & 88.8\std{3.4} & 99.6\std{0.7} & 83.1\std{2.1} & 84.9\std{3.7} & 85.5\std{1.3} & \textbf{85.8\std{1.5}} & 67.6\std{2.2} \\
RL & HGPO & 78.3\std{7.9} & 83.6\std{2.5} & 89.2\std{2.8} & 97.2\std{1.3} & 86.5\std{1.9} & \textbf{99.1\std{1.5}} & 87.8\std{1.3} & 83.2\std{1.6} & 65.6\std{2.5} \\
RL & ProGPO & \textbf{79.3\std{5.3}} & \textbf{86.9\std{2.5}} & \textbf{98.0\std{1.7}} & \textbf{100.0\std{0.0}} & \textbf{94.9\std{0.8}} & 84.6\std{1.1} & \textbf{90.1\std{1.1}} & 85.4\std{1.5} & \textbf{71.5\std{3.0}} \\
\bottomrule
\end{tabular*}
\renewcommand{\arraystretch}{1.0}
\end{table}

Table~\ref{tab:main_results} reports prompt-based and RL-based baselines in one comparison. Since the prompting, PPO, and RLOO entries are taken from prior reports, we use them as reference points rather than matched-protocol comparisons. Among our local reruns, ProGPO improves ALFWorld overall success from 75.8\% for GRPO, 85.5\% for GiGPO, and 87.8\% for HGPO to 90.1\%. It also leads the Pick, Clean, Cool, Look, and Heat diagnostics, while HGPO remains higher on the ALFWorld two-object task. On WebShop, ProGPO obtains the highest success rate at 71.5\%, while GiGPO is slightly higher on task score.

\begin{figure}[!htbp]
\centering
\includegraphics[width=\linewidth]{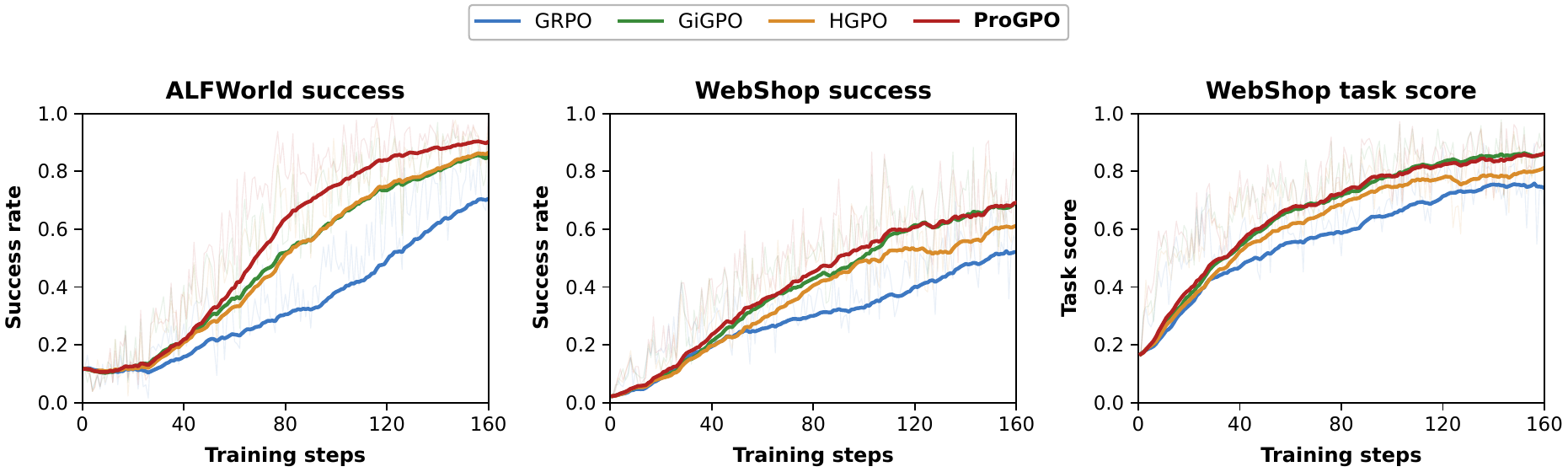}
\caption{Training curves for ProGPO (red), HGPO (orange), GiGPO (green), and GRPO (blue). Lighter curves show raw values; darker curves use EMA smoothing with decay $\alpha=0.95$.}
\label{fig:training_curves}
\end{figure}

Beyond aggregate success, Table~\ref{tab:main_results} reports the full ALFWorld task taxonomy. ProGPO is higher on Pick, Clean, Cool, Look, Heat, and All, while HGPO remains higher on Pick2. On WebShop, ProGPO has the highest success rate and GiGPO is marginally higher on task score, suggesting that ProGPO improves terminal completion more clearly than partial task progress in this setting. The larger-scale ALFWorld runs show a similar trend: at the last common non-empty checkpoint (step 155), ProGPO reaches 93.0\% training success, 94.5\% validation success, and 7.57 validation test score, compared with 89.1\%, 93.0\%, and 6.99 for GiGPO, and 86.7\%, 89.1\%, and 5.84 for HGPO.

\subsection{Mechanism Diagnostics}

\begin{figure}[!htbp]
\centering
\includegraphics[width=\linewidth]{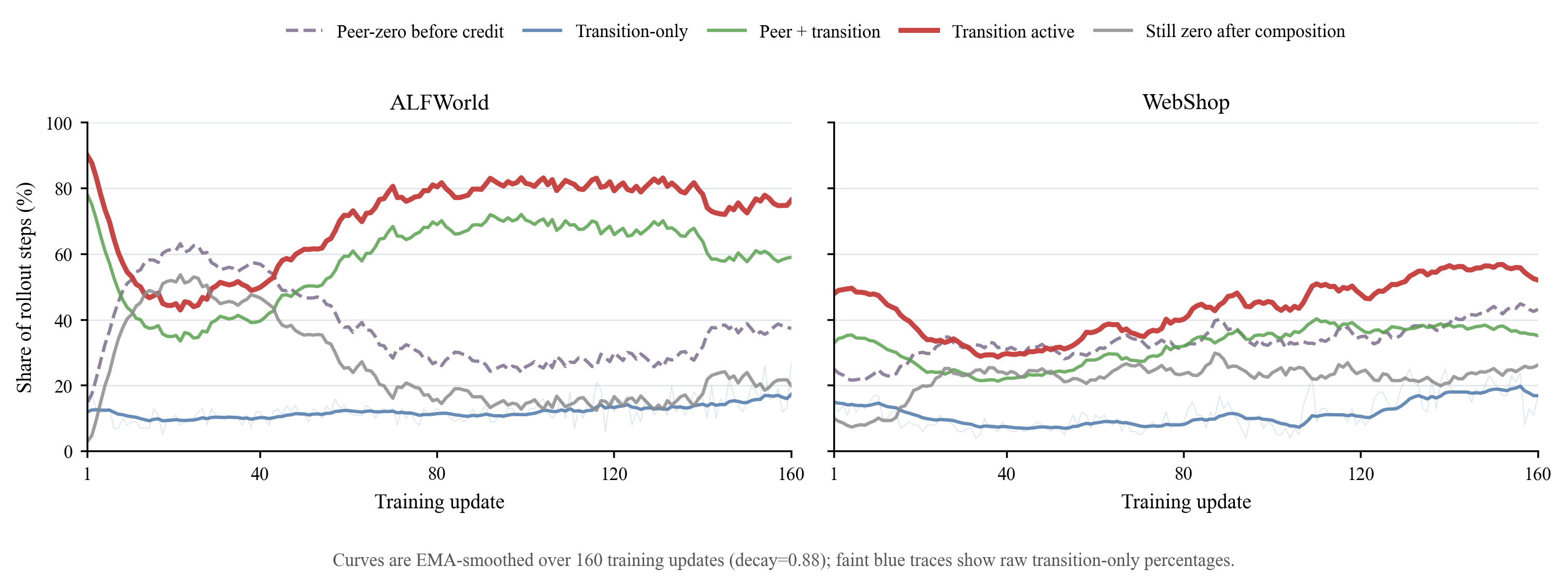}
\caption{Mechanism diagnostics over 160 training updates. Transition-active denotes the union of transition-only and peer-and-transition steps. Curves are EMA-smoothed; faint traces show raw transition-only shares.}
\label{fig:mechanism_curves}
\end{figure}

\begin{table}[H]
\centering
\caption{Mechanism diagnostics averaged over training traces. Peer-zero is measured before transition credit; still-zero is measured after credit composition.}
\label{tab:diagnostics}
\vspace{0.45em}
\footnotesize
\setlength{\tabcolsep}{6.0pt}
\renewcommand{\arraystretch}{1.18}
\begin{tabular*}{0.72\linewidth}{@{}l@{\extracolsep{\fill}}cc@{}}
\toprule
Diagnostic & WebShop & ALFWorld \\
\midrule
Peer-zero steps & 33.7\% & 44.5\% \\
Nonzero steps & 77.7\% & 66.9\% \\
Transition-only steps & 11.4\% & 11.4\% \\
Peer and transition & 31.6\% & 51.9\% \\
Transition-active steps & 43.0\% & 63.3\% \\
Still zero after composition & 22.3\% & 33.1\% \\
Higher-$k$ lookups & 82.5\% & 73.3\% \\
Active levels & 2.80 & 2.61 \\
Higher-$k$ share & 0.61 & 0.48 \\
\bottomrule
\end{tabular*}
\renewcommand{\arraystretch}{1.0}
\end{table}

Figure~\ref{fig:mechanism_curves} and Table~\ref{tab:diagnostics} show that transition credit and peer comparison cover different parts of training. Transition-only signal appears on 11.4\% of steps in both environments, but transition credit is active more broadly because it also appears on steps that already have peer-comparison signal. We call this union \emph{transition-active}; it averages 43.0\% of WebShop steps and 63.3\% of ALFWorld steps in Table~\ref{tab:diagnostics}. Transition credit is inactive when the current or next state has no reliable potential lookup, when the step has no next state, or when $\hat{V}(s_{t+1})-\hat{V}(s_t)$ is numerically zero. ProGPO therefore complements prefix-consistent peer comparison rather than converting every peer-zero step into a nonzero update. Multi-resolution fusion is used in most value lookups, with higher-depth estimates active in 82.5\% of WebShop lookups and 73.3\% of ALFWorld lookups.

\subsection{Transition-Credit Ablations}
\label{sec:ablations}

We ablate the components used to estimate transition credit on ALFWorld. These runs are training-side diagnostics at the final training update, intended to isolate the estimator components rather than replace the matched validation comparison in Table~\ref{tab:main_results}. Since semantic expansion and reliability-aware fusion both operate through the state-potential estimator, Table~\ref{tab:ablations} compares plain transition credit, each stabilization component alone, and the full ProGPO estimator.

\begin{table}[H]
\centering
\caption{ALFWorld transition-credit ablations measured on training rollouts at step 160. Success-rate columns are percentages.}
\label{tab:ablations}
\vspace{0.45em}
\small
\setlength{\tabcolsep}{4.0pt}
\renewcommand{\arraystretch}{1.18}
\begin{tabular*}{\linewidth}{@{}l@{\extracolsep{\fill}}ccccc@{}}
\toprule
Configuration & Score & All & Cool & Heat & Pick2 \\
\midrule
Transition credit only & 5.49 & 87.5 & 94.7 & 95.0 & 52.6 \\
+ semantic expansion & 6.93 & 92.2 & 94.7 & \textbf{100.0} & 78.9 \\
+ reliability fusion & 6.69 & 90.6 & 94.7 & 95.0 & 73.7 \\
+ semantic expansion + reliability fusion & \textbf{8.47} & \textbf{96.9} & 94.7 & \textbf{100.0} & \textbf{84.2} \\
\bottomrule
\end{tabular*}
\renewcommand{\arraystretch}{1.0}
\end{table}

The ablation results indicate that transition credit depends strongly on the quality of the state-potential estimate. Plain transition credit is weak on the two-object task. Adding semantic expansion improves training success from 87.5\% to 92.2\%, while reliability-aware fusion alone improves it to 90.6\%. Combining both components gives the best result, reaching 96.9\% success and the highest training score. These results support the design in Section~\ref{sec:transition}, where semantic expansion and multi-resolution fusion stabilize the potential estimates used by transition credit.

\section{Limitations and Future Work}

ProGPO still depends on useful state recurrence inside a rollout batch. Transition credit can fill part of the gap left by prefix-consistent peer comparison, but it does not turn every step into a supervised update; in our diagnostics, a residual set of steps remains zero after credit composition. This is expected when both the current and next states have nearly identical estimated potentials, or when neither state has enough reliable evidence in the batch.

The non-parametric value table also inherits the quality of the state representation. Exact-prefix matching is conservative for action comparison, but state-potential estimation can still be noisy when observations are mostly unique or when semantically retrieved neighbors correspond to superficially similar but behaviorally different states. The current semantic expansion uses policy hidden states as a practical similarity signal; it does not by itself prove that nearby states have similar expected return. A learned state abstraction, better calibrated uncertainty estimates, or task-specific filters for semantic expansion may reduce this failure mode.

Transition credit should also be interpreted as a batch-estimated progress signal rather than a causal effect estimator. Changes in $\hat{V}(s_{t+1})-\hat{V}(s_t)$ can reflect environment stochasticity, task difficulty, or representation error in addition to the chosen action. Systematic sweeps over rollout group size and method-specific hyperparameters such as $K_h$, $m$, $K_{\mathrm{sem}}$, $\lambda_{\mathrm{tr}}$, $\lambda_{\mathrm{fb}}$, and the variance floor are left for future work.

Finally, the experiments focus on ALFWorld and WebShop with Qwen2.5 policies up to 3B parameters. These environments expose long-horizon interaction and sparse rewards, but they do not cover all agent settings. Evaluating ProGPO on more open-ended web tasks, larger models, and multimodal agents would give a clearer picture of when progress-based credit is most useful.

\section{Conclusion}

This paper studied credit sparsity in context-consistent step-level group RL. Exact context matching yields fairer action comparisons, but also leaves many steps without peer-comparison signal. ProGPO addresses this limitation with rollout-derived state potentials, transition credit, and inverse-variance multi-resolution fusion across history depths. Matched WebShop and ALFWorld traces show that progress- and reliability-oriented group policy optimization improves context-consistent training without additional reward models or parametric value networks.

\bibliographystyle{plainnat}
\bibliography{references}

@article{yao2023react,
  title = {ReAct: Synergizing Reasoning and Acting in Language Models},
  author = {Yao, Shunyu and Zhao, Jeffrey and Yu, Dian and Du, Nan and Shafran, Izhak and Narasimhan, Karthik and Cao, Yuan},
  journal = {arXiv preprint arXiv:2210.03629},
  year = {2023}
}

@article{liu2024agentbench,
  title = {AgentBench: Evaluating LLMs as Agents},
  author = {Liu, Xiao and Yu, Hao and Zhang, Hanchen and Xu, Yifan and Lei, Xuanyu and Lai, Hanyu and Gu, Yu and Ding, Hangliang and Men, Kai and Yang, Kejuan and others},
  journal = {arXiv preprint arXiv:2308.03688},
  year = {2024}
}

@inproceedings{brown2020language,
  title = {Language Models are Few-Shot Learners},
  author = {Brown, Tom B. and Mann, Benjamin and Ryder, Nick and Subbiah, Melanie and Kaplan, Jared and Dhariwal, Prafulla and Neelakantan, Arvind and Shyam, Pranav and Sastry, Girish and Askell, Amanda and others},
  booktitle = {Advances in Neural Information Processing Systems},
  year = {2020}
}

@inproceedings{shridhar2021alfworld,
  title = {ALFWorld: Aligning Text and Embodied Environments for Interactive Learning},
  author = {Shridhar, Mohit and Yuan, Xingdi and Côté, Marc-Alexandre and Bisk, Yonatan and Trischler, Adam and Hausknecht, Matthew},
  booktitle = {International Conference on Learning Representations},
  year = {2021}
}

@inproceedings{yao2022webshop,
  title = {WebShop: Towards Scalable Real-World Web Interaction with Grounded Language Agents},
  author = {Yao, Shunyu and Chen, Howard and Yang, John and Narasimhan, Karthik},
  booktitle = {Advances in Neural Information Processing Systems},
  year = {2022}
}

@inproceedings{zheng2024seeact,
  title = {GPT-4V(ision) is a Generalist Web Agent, if Grounded},
  author = {Zheng, Boyuan and Gou, Boyu and Kil, Jihyung and Sun, Huan and Su, Yu},
  booktitle = {International Conference on Learning Representations},
  year = {2024}
}

@inproceedings{deng2023mind2web,
  title = {Mind2Web: Towards a Generalist Agent for the Web},
  author = {Deng, Xiang and Gu, Yu and Zheng, Boyuan and Chen, Shijie and Stevens, Sam and Wang, Boshi and Sun, Huan and Su, Yu},
  booktitle = {Advances in Neural Information Processing Systems},
  year = {2023}
}

@inproceedings{zhou2023webarena,
  title = {WebArena: A Realistic Web Environment for Building Autonomous Agents},
  author = {Zhou, Shuyan and Xu, Frank F. and Zhu, Hao and Zhou, Xuhui and Lo, Robert and Sridhar, Abishek and Cheng, Xianyi and Ou, Tianyue and Bisk, Yonatan and Fried, Daniel and Alon, Uri and Neubig, Graham},
  booktitle = {International Conference on Learning Representations},
  year = {2024}
}

@inproceedings{koh2024visualwebarena,
  title = {VisualWebArena: Evaluating Multimodal Agents on Realistic Visual Web Tasks},
  author = {Koh, Jing Yu and Lo, Robert and Jang, Lawrence and Duvvur, Vikram and Lim, Ming Chong and Huang, Po-Yao and Neubig, Graham and Zhou, Shuyan and Salakhutdinov, Ruslan and Fried, Daniel},
  booktitle = {Annual Meeting of the Association for Computational Linguistics},
  year = {2024}
}

@article{wang2024voyager,
  title = {Voyager: An Open-Ended Embodied Agent with Large Language Models},
  author = {Wang, Guanzhi and Xie, Yuqi and Jiang, Yunfan and Mandlekar, Ajay and Xiao, Chaowei and Zhu, Yuke and Fan, Linxi and Anandkumar, Anima},
  journal = {Transactions on Machine Learning Research},
  year = {2024}
}

@article{ahn2022saycan,
  title = {Do As I Can, Not As I Say: Grounding Language in Robotic Affordances},
  author = {Ahn, Michael and Brohan, Anthony and Brown, Noah and Chebotar, Yevgen and Cortes, Omar and David, Byron and Finn, Chelsea and Fu, Chuyuan and Gopalakrishnan, Keerthana and Hausman, Karol and others},
  journal = {arXiv preprint arXiv:2204.01691},
  year = {2022}
}

@inproceedings{driess2023palme,
  title = {PaLM-E: An Embodied Multimodal Language Model},
  author = {Driess, Danny and Xia, Fei and Sajjadi, Mehdi S. M. and Lynch, Corey and Chowdhery, Aakanksha and Ichter, Brian and Wahid, Ayzaan and Tompson, Jonathan and Vuong, Quan and Yu, Tianhe and others},
  booktitle = {International Conference on Machine Learning},
  year = {2023}
}

@inproceedings{brohan2023rt2,
  title = {RT-2: Vision-Language-Action Models Transfer Web Knowledge to Robotic Control},
  author = {Brohan, Anthony and Brown, Noah and Carbajal, Justice and Chebotar, Yevgen and Dabis, Joseph and Finn, Chelsea and Gopalakrishnan, Keerthana and Hausman, Karol and Herzog, Alex and Hsu, Jasmine and others},
  booktitle = {Conference on Robot Learning},
  year = {2023}
}

@inproceedings{furuta2024multimodal,
  title = {Multimodal Web Navigation with Instruction-Finetuned Foundation Models},
  author = {Furuta, Hiroki and Lee, Kuang-Huei and Nachum, Ofir and Matsuo, Yutaka and Faust, Aleksandra and Gu, Shixiang Shane and Gur, Izzeddin},
  booktitle = {International Conference on Learning Representations},
  year = {2024}
}

@inproceedings{bai2024digirl,
  title = {DigiRL: Training In-the-Wild Device-Control Agents with Autonomous Reinforcement Learning},
  author = {Bai, Hao and Zhou, Yifei and Pan, Jiayi and Cemri, Mert and Suhr, Alane and Levine, Sergey and Kumar, Aviral},
  booktitle = {Advances in Neural Information Processing Systems},
  year = {2024}
}

@inproceedings{ouyang2022training,
  title = {Training Language Models to Follow Instructions with Human Feedback},
  author = {Ouyang, Long and Wu, Jeffrey and Jiang, Xu and Almeida, Diogo and Wainwright, Carroll and Mishkin, Pamela and Zhang, Chong and Agarwal, Sandhini and Slama, Katarina and Ray, Alex and others},
  booktitle = {Advances in Neural Information Processing Systems},
  year = {2022}
}

@inproceedings{christiano2017deep,
  title = {Deep Reinforcement Learning from Human Preferences},
  author = {Christiano, Paul F. and Leike, Jan and Brown, Tom B. and Martic, Miljan and Legg, Shane and Amodei, Dario},
  booktitle = {Advances in Neural Information Processing Systems},
  year = {2017}
}

@article{bai2022training,
  title = {Training a Helpful and Harmless Assistant with Reinforcement Learning from Human Feedback},
  author = {Bai, Yuntao and Jones, Andy and Ndousse, Kamal and Askell, Amanda and Chen, Anna and DasSarma, Nova and Drain, Dawn and Fort, Stanislav and Ganguli, Deep and Henighan, Tom and others},
  journal = {arXiv preprint arXiv:2204.05862},
  year = {2022}
}

@inproceedings{rafailov2023direct,
  title = {Direct Preference Optimization: Your Language Model is Secretly a Reward Model},
  author = {Rafailov, Rafael and Sharma, Archit and Mitchell, Eric and Ermon, Stefano and Manning, Christopher D. and Finn, Chelsea},
  booktitle = {Advances in Neural Information Processing Systems},
  year = {2023}
}

@article{guo2025deepseek,
  title = {DeepSeek-R1: Incentivizing Reasoning Capability in LLMs via Reinforcement Learning},
  author = {{DeepSeek-AI} and Guo, Daya and Yang, Dejian and Zhang, Haowei and Song, Junxiao and Wang, Peiyi and Zhu, Qihao and Xu, Runxin and Zhang, Ruoyu and Ma, Shirong and others},
  journal = {arXiv preprint arXiv:2501.12948},
  year = {2025}
}

@inproceedings{kool2019rloo,
  title = {Buy 4 REINFORCE Samples, Get a Baseline for Free},
  author = {Kool, Wouter and van Hoof, Herke and Welling, Max},
  booktitle = {Deep Reinforcement Learning Meets Structured Prediction Workshop at ICLR},
  year = {2019}
}

@article{shao2024deepseekmath,
  title = {DeepSeekMath: Pushing the Limits of Mathematical Reasoning in Open Language Models},
  author = {Shao, Zhihong and Wang, Peiyi and Zhu, Qihao and Xu, Runxin and Song, Junxiao and Bi, Xiao and Zhang, Haowei and Zhang, Mingchuan and Li, Y. K. and Wu, Y. and Guo, Daya},
  journal = {arXiv preprint arXiv:2402.03300},
  year = {2024}
}

@article{yu2025dapo,
  title = {DAPO: An Open-Source LLM Reinforcement Learning System at Scale},
  author = {Yu, Qiying and Zhang, Zheng and Zhu, Ruofei and Yuan, Yufeng and Zuo, Xiaochen and Yue, Yu and Dai, Weinan and Fan, Tiantian and Liu, Gaohong and Liu, Lingjun and others},
  journal = {arXiv preprint arXiv:2503.14476},
  year = {2025}
}

@article{schulman2017proximal,
  title = {Proximal Policy Optimization Algorithms},
  author = {Schulman, John and Wolski, Filip and Dhariwal, Prafulla and Radford, Alec and Klimov, Oleg},
  journal = {arXiv preprint arXiv:1707.06347},
  year = {2017}
}

@article{williams1992simple,
  title = {Simple Statistical Gradient-Following Algorithms for Connectionist Reinforcement Learning},
  author = {Williams, Ronald J.},
  journal = {Machine Learning},
  volume = {8},
  pages = {229--256},
  year = {1992}
}

@inproceedings{sutton1999policy,
  title = {Policy Gradient Methods for Reinforcement Learning with Function Approximation},
  author = {Sutton, Richard S. and McAllester, David and Singh, Satinder and Mansour, Yishay},
  booktitle = {Advances in Neural Information Processing Systems},
  year = {1999}
}

@inproceedings{konda1999actorcritic,
  title = {Actor-Critic Algorithms},
  author = {Konda, Vijay R. and Tsitsiklis, John N.},
  booktitle = {Advances in Neural Information Processing Systems},
  year = {1999}
}

@article{peng2019advantage,
  title = {Advantage-Weighted Regression: Simple and Scalable Off-Policy Reinforcement Learning},
  author = {Peng, Xue Bin and Kumar, Aviral and Zhang, Grace and Levine, Sergey},
  journal = {arXiv preprint arXiv:1910.00177},
  year = {2019}
}

@article{cobbe2021training,
  title = {Training Verifiers to Solve Math Word Problems},
  author = {Cobbe, Karl and Kosaraju, Vineet and Bavarian, Mohammad and Chen, Mark and Jun, Heewoo and Kaiser, Lukasz and Plappert, Matthias and Tworek, Jerry and Hilton, Jacob and Nakano, Reiichiro and others},
  journal = {arXiv preprint arXiv:2110.14168},
  year = {2021}
}

@inproceedings{lightman2024verify,
  title = {Let's Verify Step by Step},
  author = {Lightman, Hunter and Kosaraju, Vineet and Burda, Yura and Edwards, Harrison and Baker, Bowen and Lee, Teddy and Leike, Jan and Schulman, John and Sutskever, Ilya and Cobbe, Karl},
  booktitle = {International Conference on Learning Representations},
  year = {2024}
}

@article{wang2025ragen,
  title = {RAGEN: Understanding Self-Evolution in LLM Agents via Multi-Turn Reinforcement Learning},
  author = {Wang, Zihan and Wang, Kangrui and Wang, Qineng and Zhang, Pingyue and Li, Linjie and Yang, Zhengyuan and Jin, Xing and Yu, Kefan and Nguyen, Minh Nhat and Liu, Licheng and others},
  journal = {arXiv preprint arXiv:2504.20073},
  year = {2025}
}

@article{jin2025searchr1,
  title = {Search-R1: Training LLMs to Reason and Leverage Search Engines with Reinforcement Learning},
  author = {Jin, Bowen and Zeng, Hansi and Yue, Zhenrui and Yoon, Jinsung and Arik, Sercan and Wang, Dong and Zamani, Hamed and Han, Jiawei},
  journal = {arXiv preprint arXiv:2503.09516},
  year = {2025}
}

@inproceedings{schick2023toolformer,
  title = {Toolformer: Language Models Can Teach Themselves to Use Tools},
  author = {Schick, Timo and Dwivedi-Yu, Jane and Dess{\`i}, Roberto and Raileanu, Roberta and Lomeli, Maria and Hambro, Eric and Zettlemoyer, Luke and Cancedda, Nicola and Scialom, Thomas},
  booktitle = {Advances in Neural Information Processing Systems},
  year = {2023}
}

@inproceedings{qin2024toolllm,
  title = {ToolLLM: Facilitating Large Language Models to Master 16000+ Real-World APIs},
  author = {Qin, Yujia and Liang, Shengding and Ye, Yining and Zhu, Kunlun and Yan, Lan and Lu, Yaxi and Lin, Yankai and Cong, Xin and Tang, Xiangru and Qian, Bill and others},
  booktitle = {International Conference on Learning Representations},
  year = {2024}
}

@article{feng2025gigpo,
  title = {Group-in-Group Policy Optimization for LLM Agent Training},
  author = {Feng, Lang and Xue, Zhenghai and Liu, Tingcong and An, Bo},
  journal = {arXiv preprint arXiv:2505.10978},
  year = {2025}
}

@inproceedings{zhang2025hgpo,
  title = {Hierarchy-of-Groups Policy Optimization for Long-Horizon Agentic Tasks},
  author = {He, Shuo and Feng, Lang and Wei, Qi and Cheng, Xin and Feng, Lei and An, Bo},
  booktitle = {International Conference on Learning Representations},
  year = {2026}
}

@inproceedings{shinn2023reflexion,
  title = {Reflexion: Language Agents with Verbal Reinforcement Learning},
  author = {Shinn, Noah and Cassano, Federico and Berman, Edward and Gopinath, Ashwin and Narasimhan, Karthik and Yao, Shunyu},
  booktitle = {Advances in Neural Information Processing Systems},
  year = {2023}
}

\end{document}